\documentclass[letterpaper, 10 pt, conference]{ieeeconf}  % Comment this line out if you need a4paper

\IEEEoverridecommandlockouts                              % This command is only needed if 
\usepackage{amsmath} 
\usepackage{amssymb}  
\usepackage{bm}
\usepackage{cite}
\usepackage{placeins}

\usepackage{graphicx}

\title{\LARGE \bf
Towards Efficient Robotic Manipulation Models with Self-Recursive Pruning
}

\author{
  Zijia Chen$^{1}$ \quad
  Yuenan Hou$^{2,\dagger}$ \quad
  Yu Li$^{2}$ \quad
  Weijie Li$^{1}$ \quad
  Li Liu$^{1,\dagger}$ \\
  $^{1}$College of Electronic Science and Technology, National University of Defense Technology  \\
  Changsha, 410073, China \\
  $^{2}$Shanghai AI Laboratory \\
  Shanghai, 200000, China \\
  {\small $^\dagger$Corresponding authors.}
}

\begin{document}

\maketitle
\thispagestyle{empty}
\pagestyle{empty}

%%%%%%%%%%%%%%%%%%%%%%%%%%%%%%%%%%%%%%%%%%%%%%%%%%%%%%%%%%%%%%%%%%%%%%%%%%%%%%%%
\begin{abstract}

Network pruning can reduce parameter redundancy in robotic policies. However, generic pruning criteria are tailored for image recognition tasks and commonly designed to preserve weight magnitude, local reconstruction, or language-model likelihood rather than closed-loop action behavior. Directly applying these pruning algorithms to robotic tasks yields unsatisfactory performance. In this paper, we propose Loss-Conditioned Activation-Moment (LCAM) pruning, a training-free method for unstructured pruning of pre-trained robotic manipulation policies. Specifically, we first rank connections using row-normalized weight contribution, activation moments measured on calibration demonstrations, and the sensitivity of output directions to the action-prediction loss. We further design a self-recursive coarse-to-fine procedure: importance is recalibrated after each nested coarse pruning stage, while held-out offline action distortion guides fine-grained budget allocation after a sparsity knee. Our algorithm is free from costly recovery training and simulator rollouts after pruning. Experiments on three LIBERO suites with competitive robotic policies, together with evaluations on OpenVLA, show that LCAM attains competitive performance across a broad range of pruning ratios. Notably, on LIBERO-Object with OpenVLA, our LCAM achieves $84.0\%$ success at $50\%$ unstructured pruning, retaining over $90\%$ of the dense policy's success rate. Promising results on real-world robotic ping pong further demonstrate the effectiveness of our pruning algorithm. %Ablation results further demonstrate the importance of the self-recursive pruning procedure and the components of the LCAM importance criterion.

\end{abstract}

%%%%%%%%%%%%%%%%%%%%%%%%%%%%%%%%%%%%%%%%%%%%%%%%%%%%%%%%%%%%%%%%%%%%%%%%%%%%%%%%
\section{INTRODUCTION}

\begin{figure*}[t]
    \centering
    \includegraphics[width=\textwidth]{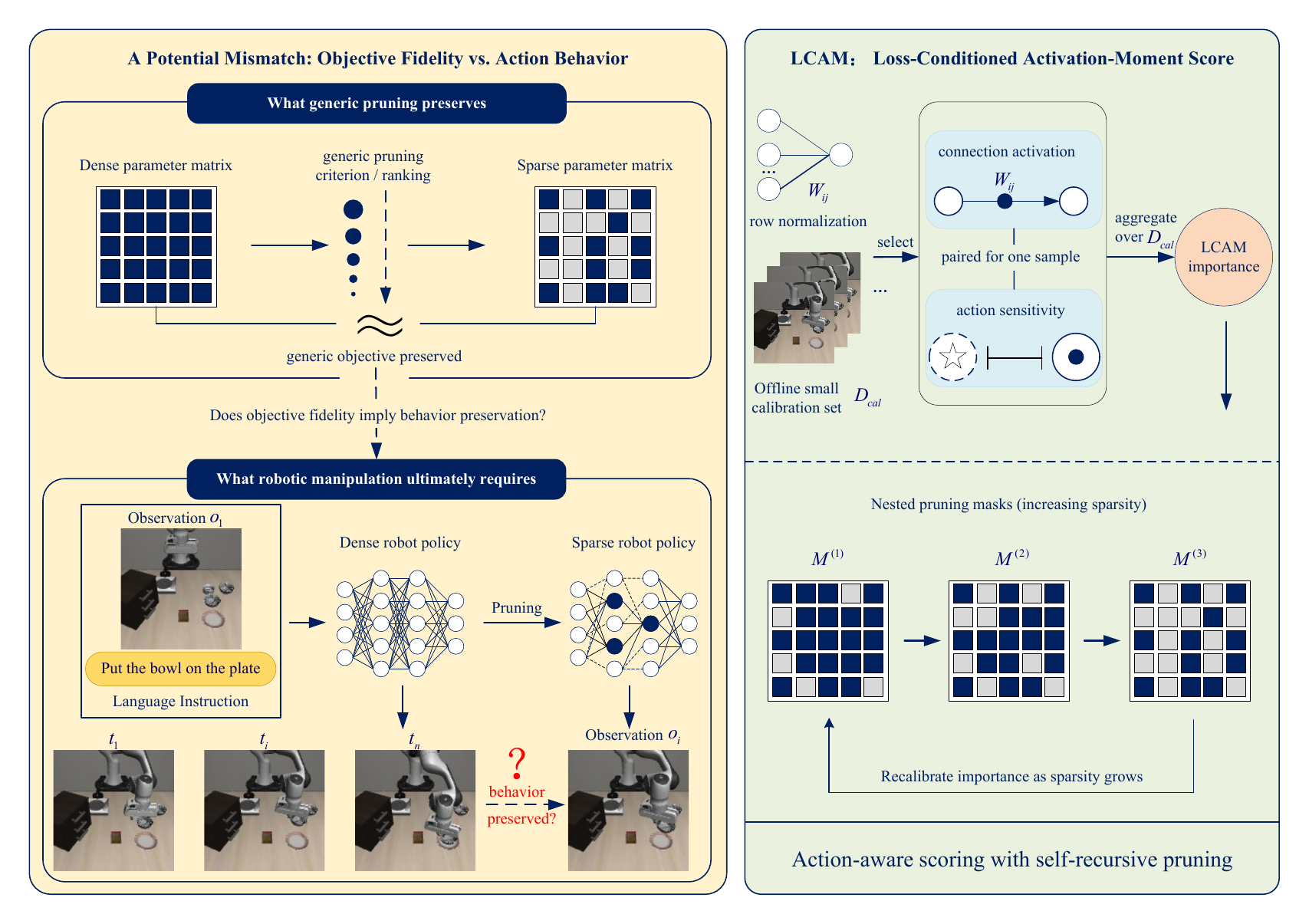}    \caption{Comparison between generic pruning and our LCAM. Generic pruning may preserve its pruning objective without directly ensuring task-relevant action behavior. LCAM evaluates each connection using its row-normalized weight contribution and action-conditioned activation evidence from a small offline calibration set, and recursively recalibrates connection importance as sparsity increases to construct nested pruning masks.}
    \label{fig:overview}
\end{figure*}

Recent advances in robot learning have produced action policies that achieve strong manipulation performance across diverse manipulation tasks~\cite{brohan2023rt1,zhao2023act,shridhar2023peract,chi2025diffusion, zitkovich2023rt2,octo2024,kim2025openvla,black2025pi0,ddpg-per}. However, modern robotic manipulation policies typically rely on high-capacity perception and cumbersome policy networks, leading to substantial storage and inference costs. This motivates the compression of robotic policies, with the objective of reducing parameter redundancy while maximally preserving the ultimate performance. We focus on unstructured pruning, which removes individual weights to induce sparsity in pretrained policies. Classical magnitude pruning and recent one-shot methods, such as SparseGPT and Wanda, have demonstrated effective unstructured sparsification for pretrained image recognition and language models~\cite{han2015learning,frantar2023sparsegpt,sun2024wanda}. These pruning methods typically employ classification accuracy, perplexity, or reconstruction errors as pruning criteria. However, directly applying these criteria to robotic manipulation policies can be challenging.

The challenge stems from the closed-loop nature of robotic manipulation. Unlike language modeling, where pruning quality can often be assessed using token-level likelihood or layer-wise reconstruction errors, robotic policies are ultimately evaluated by the task success rate during closed-loop execution. Small action deviations introduced by pruning can accumulate during closed-loop execution and thus yield degraded manipulation performance~\cite{ross2011reduction}. Moreover, robotic demonstrations are typically scarce and task-specific, while modern policies combine heterogeneous perception, state, language, and action-generation modules that differ markedly in their sensitivity to parameter removal. To mitigate such degradation, recent robotic policy compression and post-pruning recovery methods often couple structural reduction with additional optimization or test-time correction~\cite{wu2025lightdp,chen2026rlrc,jeon2026shallowpi,
jabbour2025gluestick}. These procedures either introduce additional training cost or require intervention in test-time. Consequently, pruning objectives based on generic activation statistics or local reconstruction fidelity can not fully reflect the task-relevant behavior of robotic policies under closed-loop execution.

To address the aforementioned challenges, we propose \emph{Loss-Conditioned Activation-Moment} (LCAM) pruning, a training-free framework for sparsifying pretrained robotic policies. At the connection level, LCAM ranks connections by combining row-normalized weight contribution, activation evidence on calibration demonstrations, and action-loss sensitivity. High-score connections are retained, whereas low-score connections are removed. At the allocation level, LCAM adopts a self-recursive coarse-to-fine pruning procedure. Starting from the dense policy, each subsequent stage begins with the previously pruned policy, recalibrates connection importance on the resulting sparse model, and removes only redundant connections. The procedure automatically transitions from coarse to fine-grained pruning when offline action distortion on validation demonstrations exhibits a sharp marginal increase, indicating a sparsity knee. It then refines the remaining pruning budget using smaller increments to preserve action behavior at high sparsity. Note that LCAM requires neither post-pruning retraining nor simulator rollouts during pruning.

We evaluate our LCAM against representative training-free unstructured pruning baselines on robotic manipulation policies in the LIBERO benchmark~\cite{liu2023libero}, covering three LIBERO suites with MATE~\cite{chen2026mate} policies and an evaluation on OpenVLA~\cite{kim2025openvla}. The results show that LCAM produces a competitive training-free pruning strategy across multiple sparsity levels. On the MATE policies, LCAM retains substantial task success rates in high-sparsity regimes where several weight-based, saliency-based, and reconstruction-based baselines undergo sharp performance drops or collapse. On LIBERO-Object where OpenVLA is adopted, LCAM achieves the highest success rate among competitive baselines at both $50\%$ and $60\%$ pruning, attaining $84.0\%$ success at $50\%$ pruning and retaining over $90\%$ of the dense policy's success rate. Promising results on the real-world robotic ping pong task further underscore the effectiveness of our pruning algorithm.

%%%%%%%%%%%%%%%%%%%%%%%%%%%%%%%%%%%%%%%%%%%%%%%%%%%%%%%%%%%%%%%%%%%%%%%%%%%%%%%%
\section{RELATED WORK}

\subsection{Network Pruning}

Network pruning aims to identify and remove redundant parameters while preserving the task performance of a pretrained model. Magnitude pruning is a classical and widely used baseline that removes parameters according to their absolute weight values~\cite{han2015learning,peel}. Despite its simplicity, magnitude pruning ranks parameters solely by their absolute values, without accounting for input-dependent activation statistics or action-loss sensitivity.

More recent pruning methods incorporate additional information, such as loss sensitivity, activation statistics, or layer-wise reconstruction, to estimate parameter importance. Taylor-based pruning estimates the loss change induced by removing a parameter group using first- or second-order Taylor expansions~\cite{molchanov2019importance}. Although originally developed for structured pruning, its loss-based saliency principle also motivates parameter-level importance criteria. For large pretrained Transformers, SparseGPT formulates pruning as a layer-wise reconstruction problem and uses approximate second-order information to enable accurate one-shot pruning~\cite{frantar2023sparsegpt}. Wanda provides a simpler training-free alternative by combining weight magnitude with input activation statistics, without updating the retained weights~\cite{sun2024wanda}. Pruner-Zero further uses genetic programming to search symbolic pruning metrics over weights, activations, gradients, and mathematical
operators~\cite{dong2024prunerzero}. AutoPrune takes a different approach by using an LLM-guided search procedure to automatically discover pruning metrics, together with a skew-aware dynamic sparsity allocation strategy for large language models~\cite{kang2026autoprune}.

These methods were primarily developed and evaluated for image classification or language modeling, where pruning is commonly guided by proxy quantities such as loss sensitivity, activation statistics, or layer-wise reconstruction, and evaluated using task-level metrics such as classification accuracy or perplexity. Whether these criteria can reliably preserve task-relevant action behavior when transferred to robotic action policies remains an empirical question, since the optimization objectives and evaluation protocols differ substantially across these settings.

\subsection{Model Compression and Pruning for Robotic Policies}

Recent work has explored the compression of robotic policies. LightDP compresses diffusion policies through a unified pruning-and-retraining pipeline and combines network compression with consistency distillation to reduce the number of denoising steps~\cite{wu2025lightdp}. RLRC applies structured pruning to vision-language-action models and recovers performance through supervised fine-tuning and reinforcement learning~\cite{chen2026rlrc}. Shallow-$\pi$ reduces the depth of flow-based VLA models and trains the compressed model through knowledge distillation~\cite{jeon2026shallowpi}. These approaches couple structural compression with additional optimization or distillation to preserve policy performance.

Recent work has also investigated post-pruning recovery without additional training. GLUESTICK recovers a structurally pruned vision-language-action model by approximating the weight difference between the dense and pruned models with low-rank components, which are added as layer-wise corrections during inference~\cite{jabbour2025gluestick}. Unlike these approaches, LCAM constructs pruning masks without recovery training, distillation, or inference-time correction.

%%%%%%%%%%%%%%%%%%%%%%%%%%%%%%%%%%%%%%%%%%%%%%%%%%%%%%%%%%%%%%%%%%%%%%%%%%%%%%%%

\section{METHOD}
\label{sec:method}

\begin{figure*}[t]
    \centering
    \includegraphics[width=\textwidth]{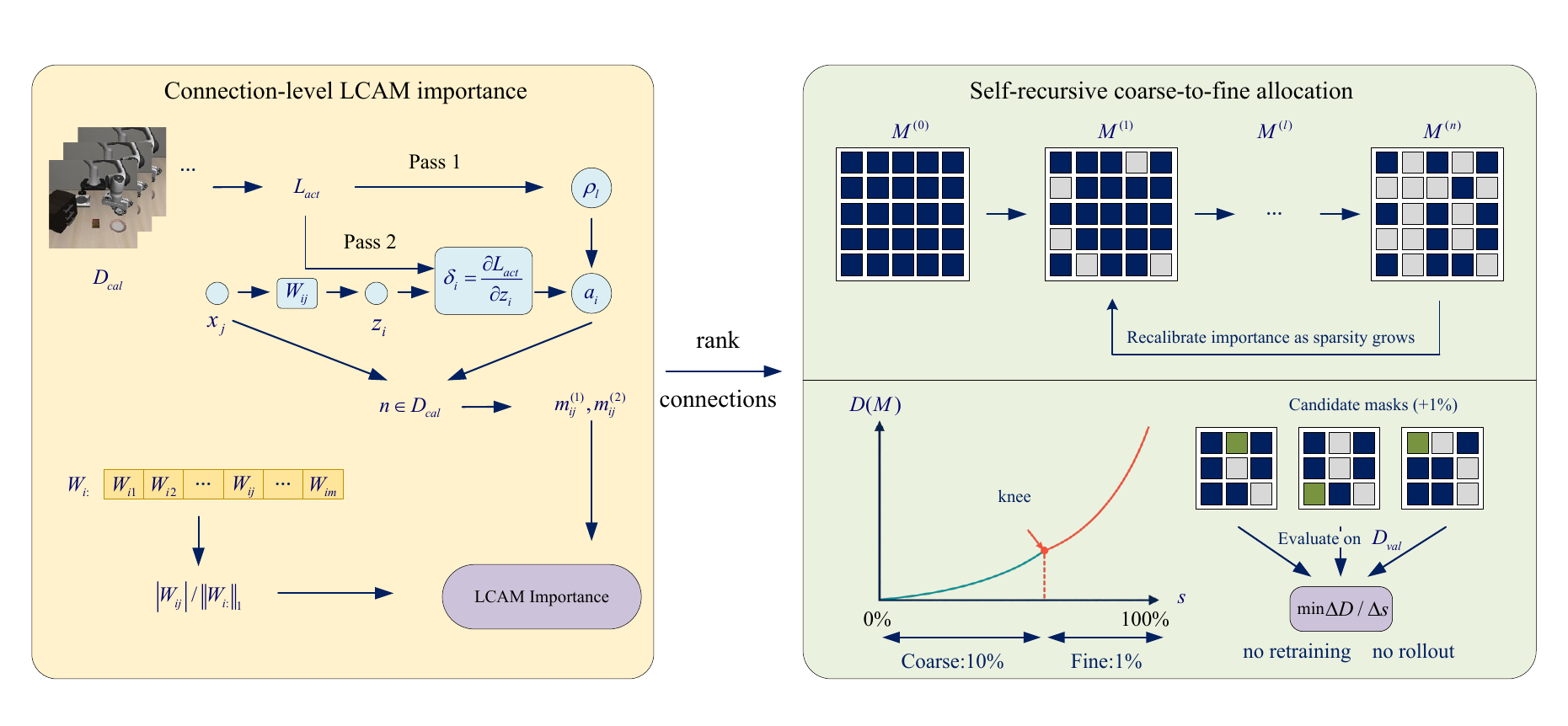}
    \caption{Framework overview of LCAM with self-recursive coarse-to-fine pruning. 
    \textbf{Left:} a first calibration pass estimates the layer-wise gradient scale, while a second pass combines row-normalized weight contribution with loss-conditioned activation moments to score individual connections. 
    \textbf{Right:} connection importance is recalibrated as nested masks become sparser. Offline action distortion \(D(M)\), evaluated on the held-out set \(\mathcal{D}_{\mathrm{val}}\), identifies a sparsity knee; afterward, equal-size candidate masks that prune an additional \(1\%\) of eligible weights are compared across feasible functional groups, and the candidate with the smallest distortion increase per newly pruned connection is accepted. LCAM constructs these masks without retraining or simulator rollouts.}
    \label{fig:method-overview}
\end{figure*}

We consider post-training unstructured pruning of a pretrained robotic action policy. LCAM consists of a connection-level importance criterion and a self-recursive coarse-to-fine allocation procedure. Section~\ref{subsec:lcam} defines the LCAM score for ranking individual connections, while Section~\ref{subsec:coarse_to_fine} presents the validation-controlled coarse-to-fine procedure that allocates a target sparsity budget without retraining or rollout-based mask selection.

\subsection{Loss-Conditioned Activation-Moment Importance}
\label{subsec:lcam}

Let $\pi_{\theta}$ denote a pretrained robotic action policy that maps an
observation--language input $\mathbf{o}$ to an action prediction
$\hat{\mathbf{a}}=\pi_{\theta}(\mathbf{o})$. Let
$\mathcal{L}_{\mathrm{act}}$ denote the native differentiable
action-prediction objective used to train $\pi_{\theta}$. During LCAM
calibration, we evaluate this objective on calibration demonstrations
solely to obtain the action-loss gradients used for importance estimation.
LCAM introduces no separate pruning loss and does not optimize
$\mathcal{L}_{\mathrm{act}}$ during mask construction. The
model-specific instantiations of $\mathcal{L}_{\mathrm{act}}$ used for
MATE and OpenVLA are given in the implementation appendix. Given a target
eligible-weight sparsity $s$, our goal is to find a binary mask
$\mathbf{M}$ that removes redundant connections while preserving the
action-prediction behavior of the dense policy.

For a linear layer $\mathbf{z}=\mathbf{W}\mathbf{x}$, pruning connection $W_{ij}$ changes its value by $\Delta W_{ij}=-W_{ij}$. A first-order Taylor approximation relates its pruning sensitivity to $\left|W_{ij}(\partial\mathcal{L}_{\mathrm{act}}/\partial z_i)x_j\right|$. This indicates that an important connection should have non-negligible relative weight, receive informative activations, and affect an action-sensitive output direction.

Let $\mathcal{D}_{\mathrm{cal}}$ denote a fixed calibration set of demonstrations, disjoint from the held-out allocation set $\mathcal{D}_{\mathrm{val}}$ used later for sparsity allocation. For each linear layer $\ell$, we collect statistics over the inputs and action-loss gradients induced by $\mathcal{D}_{\mathrm{cal}}$. Here,
$n$ indexes an action-supervised calibration instance, namely an action-prediction timestep for continuous-action policies or a predicted action-token position for token-based policies, whereas $i$ and $j$ index the output and input channels of the layer, respectively. Let $\delta_i^{(n)} = \partial\mathcal{L}_{\mathrm{act}}/\partial z_i^{(n)}$ denote the gradient of the action-prediction loss with respect to output component $z_i$ for index $n$.

LCAM estimates these factors using a two-pass calibration procedure. In the first pass, we compute the layer-wise gradient scale $\rho_{\ell} = \sqrt{\max\!\left( \mathbb{E}_{n,i}\left[(\delta_i^{(n)})^2\right],\epsilon \right)}$, where $\mathbb{E}_{n,i}$ denotes averaging over calibration indices and output channels, and $\epsilon>0$ is a numerical stability constant. In the second pass, the normalized gradient factor $a_i^{(n)}=\sqrt{1+(\delta_i^{(n)}/\rho_{\ell})^2}$ modulates the input activation observed at the same calibration index. We then collect the joint activation moments:

\begin{equation}
\begin{aligned}
m_{ij}^{(1)}
&=
\mathbb{E}_{n}
\left[
a_i^{(n)}
\left|x_j^{(n)}\right|
\right],\\
m_{ij}^{(2)}
&=
\mathbb{E}_{n}
\left[
\left(
a_i^{(n)}x_j^{(n)}
\right)^2
\right].
\end{aligned}
\label{eq:joint_moments}
\end{equation}
These moments measure input activations weighted by action-loss sensitivity at the same calibration instances. They therefore retain the sample-wise coupling between activation evidence and action-loss sensitivity.

We define the Loss-Conditioned Activation-Moment (LCAM) score as:
\begin{equation}
S_{ij}^{\mathrm{LCAM}}
=
\frac{|W_{ij}|}
{\sum_k |W_{ik}|+\epsilon}
\sqrt{
m_{ij}^{(1)}
+
\sqrt{m_{ij}^{(2)}}
+
\epsilon
}.
\label{eq:lcam_score}
\end{equation}
The row-normalized weight contribution mitigates row-wise scale differences among output directions, while the joint moments favor connections that are repeatedly active in action-sensitive contexts. Connections with lower scores are considered less important within their layer.

For comparisons across layers or modules, we use the scale-aware native
cost
$C_{ij}^{\mathrm{LCAM}} =
S_{ij}^{\mathrm{LCAM}}\|\mathbf{W}_{i:}\|_1\rho_{\ell}$,
which rescales the row-normalized score by the output-row $\ell_1$ norm
and the layer-wise gradient scale. The pruning scope and retention
constraints are fixed before mask construction and detailed in the
appendix.

\subsection{Self-Recursive Coarse-to-Fine Allocation}
\label{subsec:coarse_to_fine}

Connection-level importance does not by itself determine how a large sparsity budget should be distributed across policy components. We therefore introduce an automatic coarse-to-fine procedure. It uses $\mathcal{D}_{\mathrm{cal}}$ for LCAM calibration and a disjoint held-out allocation set $\mathcal{D}_{\mathrm{val}}$ for offline action-distortion measurement. Simulator rollouts and task success rates are not accessed during mask construction.

For a candidate mask $\mathbf{M}$, we measure the normalized deviation from the dense policy as
\begin{equation}
D(\mathbf{M}) =
\frac{1}{|\mathcal{D}_{\mathrm{val}}| d_a}
\sum_{\mathbf{o}\in\mathcal{D}_{\mathrm{val}}}
\left\|
\frac{
\pi_{\theta\odot\mathbf{M}}(\mathbf{o})
-
\pi_{\theta}(\mathbf{o})
}{
\max(\boldsymbol{\sigma},\tau)
}
\right\|_2^2 ,
\label{eq:action_distortion}
\end{equation}
where $d_a$ is the action dimension, $\boldsymbol{\sigma}$ is the vector of per-dimension action standard deviations estimated from $\mathcal{D}_{\mathrm{cal}}$, and $\tau$ is a positive stability floor. The maximum and division are applied element-wise. For action-token policies, tokens are decoded into continuous robot actions before computing $D(\mathbf{M})$.

During the coarse stage, sparsity increases in fixed increments, e.g., $10\%$. At each accepted stage, we recalibrate LCAM on the currently sparse policy and remove only additional connections. The masks are thus nested. Let $s_t$ and $D_t=D(\mathbf{M}_t)$ denote the realized sparsity and action distortion at coarse stage $t$, respectively. We compute the marginal distortion
$c_t=(D_t-D_{t-1})/(s_t-s_{t-1})$. Let
$\mathcal{H}_{<t}=\{c_2,\ldots,c_{t-1}\}$ denote the history of marginal distortions. After collecting at least two prior marginal values, we reject the current candidate when
\begin{equation}
c_t >
\operatorname{median}(\mathcal{H}_{<t})
+
\beta\widehat{\operatorname{MAD}}(\mathcal{H}_{<t}),
\label{eq:knee}
\end{equation}
where $\beta>0$ is a fixed knee-sensitivity multiplier and $\widehat{\operatorname{MAD}}(\cdot)=1.4826\operatorname{MAD}(\cdot)$ is the scaled median absolute deviation. When the criterion is triggered, the current coarse candidate is rejected and the policy is restored to the preceding accepted mask, which defines the safe boundary for fine allocation.

Starting from the safe boundary, the fine stage allocates the remaining pruning budget in small increments, e.g., $1\%$. For each feasible functional group, we construct a candidate mask by removing its lowest-cost remaining nonzero connections. All candidates are evaluated on the same fixed $\mathcal{D}_{\mathrm{val}}$, and we permanently apply the group that yields the smallest action-distortion increase per newly pruned connection. LCAM therefore determines which connections are removed within each candidate group, while offline action distortion determines which group is selected for the next fine-grained pruning increment.

All accepted updates remove additional connections only. The resulting masks remain nested throughout both stages. The procedure performs no gradient-based parameter optimization, recovery training, distillation, simulator rollout, or test-success-guided mask selection.

%\subsection{Theoretical Analysis}

\section{EXPERIMENTS}
\label{sec:experiments}

\subsection{Experimental Setup}
\label{subsec:setup}

\paragraph{Benchmarks and Policies}
We evaluate pruning algorithms on three LIBERO suites, including LIBERO-Spatial, LIBERO-Object, and LIBERO-Goal. We use MATE as the primary robotic manipulation policy backbone~\cite{chen2026mate}. We additionally evaluate LCAM on OpenVLA, a representative open-source vision-language-action model for robotic manipulation~\cite{kim2025openvla}. Specifically, we use the official OpenVLA-7B checkpoint fine-tuned on LIBERO-Object and evaluate it on the corresponding LIBERO-Object suite. The observation modalities, language instructions, image resolution, action representation, and policy architectures follow the original training and evaluation protocols of the
respective models.

\paragraph{Data protocol}
For each backbone, we use disjoint demonstration streams for a calibration set $\mathcal{D}_{\mathrm{cal}}$ and a held-out allocation set $\mathcal{D}_{\mathrm{val}}$. The former is used to estimate LCAM importance statistics, whereas the latter is used only to measure offline action distortion during coarse-to-fine allocation. Therefore, neither test trajectories nor task success rates are used to select pruning masks, transition points, pruning allocations, or hyperparameters.

\paragraph{Pruning scope and configuration}
Unless otherwise stated, we evaluate target eligible-weight pruning ratios of $10\%$, $20\%$, $30\%$, $40\%$, $50\%$, and $60\%$. We apply unstructured pruning to both MATE and OpenVLA under their respective eligible parameter scopes. For each backbone, the eligible scope is fixed before pruning and kept identical across all compared methods. Backbone-specific pruning scopes and implementation details are provided in the appendix.

\paragraph{Implementation details}
All experiments are conducted on a single NVIDIA GeForce RTX 4090 GPU with 48\,GB of memory. Unless otherwise stated, LCAM uses 16 calibration batches and 16 held-out validation batches. The coarse and fine pruning increments are $10\%$ and $1\%$, respectively, and the knee-sensitivity multiplier is set to $\beta=2.0$.

\subsection{Baselines}
\label{subsec:baselines}

We compare LCAM with representative training-free unstructured pruning baselines. Random pruning removes eligible weights uniformly at random and serves as a lower-bound reference. Magnitude pruning ranks weights by their absolute values. Wanda combines weight magnitude with input activation statistics. Taylor pruning estimates parameter saliency using first-order action-loss sensitivity.

We further include a Pruner-Zero-style gradient-saliency baseline. Auto-Prune uses the pruning criterion automatically discovered by its original search procedure. These baselines cover random, weight-based, activation-aware, loss-sensitive, and automatically discovered pruning criteria.

For each backbone, all methods use the same eligible tensor scope and requested eligible-weight sparsity. Methods that require calibration use the same calibration budget. No method uses recovery training or fine-tuning after pruning.

\subsection{Results}
\label{subsec:main_results}

We evaluate rollout success under unstructured pruning on LIBERO-Spatial, LIBERO-Object, and LIBERO-Goal. Within each policy backbone, all methods operate on the same predefined eligible parameter scope at the same target eligible-weight sparsity. Unless otherwise stated, we evaluate pruning ratios from $10\%$ to $60\%$. The Dense row denotes the unpruned policy.

Table~\ref{tab:main_results} summarizes the MATE results. At low and moderate sparsity, several baselines can remain competitive on individual suites. For example, Taylor and Pruner-Zero retain competitive performance on
LIBERO-Object up to $40\%$ pruning. However, no compared baseline exhibits consistently robust behavior across all three suites as sparsity increases.

Random pruning collapses rapidly, reaching near-zero success after $20\%$ pruning. Magnitude pruning exhibits abrupt failures on both LIBERO-Spatial and LIBERO-Object, while Wanda and Auto-Prune retain more performance at intermediate sparsity but also deteriorate sharply in the high-sparsity regime. These results show that weight-only or activation-based criteria can be highly sensitive to the policy suite and target pruning ratio.

Gradient-based baselines are more resilient in selected settings, but their robustness is likewise suite-dependent. Taylor and Pruner-Zero preserve relatively strong performance on LIBERO-Object at moderate sparsity, but both degrade substantially as sparsity increases. In comparison, LCAM retains $65.5\%$, $68.5\%$, and $55.0\%$ success on LIBERO-Spatial, LIBERO-Object, and LIBERO-Goal, respectively, at $60\%$ pruning. This comparison indicates that LCAM provides more consistent high-sparsity retention across the three MATE suites.

We additionally observe non-monotonic success-rate curves for both LCAM and the baselines. In closed-loop manipulation, small changes in predicted actions can alter the execution trajectory, contact sequence, or outcome of an episode. Consequently, rollout success need not decrease monotonically with the unstructured pruning ratio, even though sufficiently aggressive pruning eventually degrades policy behavior.

\begin{table*}[t]
\caption{Rollout success rate (\%) under unstructured pruning
of MATE policies. The dense success rate at $0\%$ pruning is shared by all
methods within each suite.}
\label{tab:main_results}
\centering
\scriptsize
\setlength{\tabcolsep}{2.2pt}
\renewcommand{\arraystretch}{1.10}
\begin{tabular}{lcccccc|cccccc|cccccc}
\hline
& \multicolumn{6}{c|}{LIBERO-Spatial}
& \multicolumn{6}{c|}{LIBERO-Object}
& \multicolumn{6}{c}{LIBERO-Goal} \\
\hline
Dense
& \multicolumn{6}{c|}{71.5}
& \multicolumn{6}{c|}{72.0}
& \multicolumn{6}{c}{66.0} \\
\hline
Method
& $10\%$ & $20\%$ & $30\%$ & $40\%$ & $50\%$ & $60\%$
& $10\%$ & $20\%$ & $30\%$ & $40\%$ & $50\%$ & $60\%$
& $10\%$ & $20\%$ & $30\%$ & $40\%$ & $50\%$ & $60\%$ \\
\hline
Random
& 29.5 & 1.0 & 0.0 & 0.0 & 0.0 & 0.0
& 8.5 & 3.0 & 0.0 & 0.0 & 0.0 & 0.0
& 7.5 & 1.0 & 0.5 & 1.0 & 0.0 & 0.0 \\
Magnitude
& 64.0 & 60.5 & 9.0 & 6.0 & 6.5 & 0.0
& 69.5 & 70.5 & 72.5 & 27.5 & 0.0 & 0.0
& 56.5 & 36.5 & 30.0 & 18.0 & 8.0 & 3.0 \\
Wanda
& 69.5 & 60.5 & 52.0 & 46.0 & 24.0 & 14.5
& 68.0 & 72.0 & 65.5 & 50.5 & 34.0 & 0.0
& 64.0 & 59.0 & 58.5 & 48.5 & 25.5 & 30.0 \\
Taylor
& 66.5 & 66.0 & 61.5 & 56.0 & 38.0 & 11.0
& 72.5 & 70.5 & 66.5 & 62.0 & 27.5 & 1.5
& 61.0 & 63.0 & 54.0 & 33.5 & 17.5 & 2.0 \\
Pruner-Zero
& 71.5 & 63.5 & 59.5 & 62.0 & 39.5 & 24.0
& 69.5 & 72.5 & 56.0 & 63.5 & 20.0 & 3.5
& 60.0 & 57.0 & 45.0 & 42.5 & 15.0 & 17.0 \\
% SparseGPT
% & 65.0 & 67.0 & 67.0 & 56.5 & 55.5 & 39.0
% & 70.0 & 65.0 & 68.5 & 62.0 & 62.5 & 19.0
% & 61.0 & 64.5 & 62.0 & 61.0 & 48.0 & 37.5 \\
Auto-Prune
& 68.0 & 60.0 & 47.0 & 34.0 & 24.0 & 5.0
& 69.0 & 69.0 & 68.5 & 61.5 & 17.0 & 0.0
& 63.5 & 58.5 & 59.0 & 48.0 & 34.0 & 7.0 \\
\hline
\textbf{LCAM (Ours)}
& 71.0 & 64.5 & 64.0 & 69.0 & 68.0 & 65.5
& 68.5 & 66.0 & 65.5 & 72.5 & 68.5 & 68.5
& 64.5 & 65.5 & 65.5 & 61.5 & 58.0 & 55.0 \\
\hline
\end{tabular}
\end{table*}

To assess cross-backbone applicability, we additionally evaluate LCAM and the compared baselines on OpenVLA fine-tuned for LIBERO-Object. Table~\ref{tab:openvla_results} reports results from $40\%$ to $60\%$ pruning, where the compared methods exhibit substantial performance separation. LCAM achieves the best success rate at $50\%$ and $60\%$
pruning, retaining $84.0\%$ and $68.5\%$ success, respectively. In contrast, Taylor pruning collapses by $50\%$, while Magnitude degrades substantially at $60\%$.

\begin{table}[t]
\centering
\caption{High-sparsity results on OpenVLA LIBERO-Object.
Entries are rollout success rates (\%).}
\label{tab:openvla_results}
\scriptsize
\setlength{\tabcolsep}{4.0pt}
\renewcommand{\arraystretch}{1.05}
\begin{tabular}{lcccc}
\hline
Method & Dense & $40\%$ & $50\%$ & $60\%$ \\
\hline
Random       & 89.5 & 0.0  & 0.0  & 0.0 \\
Magnitude    & 89.5 & 87.5 & 74.5 & 30.5 \\
Wanda        & 89.5 & 82.5 & 77.0 & 61.0 \\
Taylor       & 89.5 & 43.0 & 3.0  & 0.0 \\
Pruner-Zero  & 89.5 & 79.5 & 76.5 & 64.5 \\
% SparseGPT    & 89.5 & 85.5 & 78.0 & 51.0 \\
Auto-Prune   & 89.5 & 83.5 & 80.5 & 64.5 \\
\hline
\textbf{LCAM (Ours)}
             & 89.5 & 84.0 & \textbf{84.0} & \textbf{68.5} \\
\hline
\end{tabular}
\end{table}

\textbf{Robotic ping pong.} We construct the robotic table-tennis system using Unitree Z1 Pro and the 2D gantry. A ball machine serves the ball, and the robot is required to catch and return the ball to the opposite side. A trial is considered successful when the robot catches and returns the ball to the opposite side. Otherwise, it is considered as failed. Across 100 trials, the pruned model using our algorithm obtains a comparable ball return rate (86\%) with the original dense model (88\%), demonstrating the effectiveness of our pruning algorithms in reducing parameter redundancy and retaining model performance.

\subsection{Ablation Study}
\label{subsec:ablation}

We conduct ablations on LIBERO-Object using the same policy checkpoint, eligible parameter scope, calibration budget, pruning seed, and rollout protocol. We separately analyze the pruning procedure and the connection-level importance criterion.

\paragraph{Effect of self-recursive coarse-to-fine pruning}
Table~\ref{tab:procedure_ablation} compares one-shot native pruning, nested coarse pruning, and the full LCAM procedure. One-shot pruning constructs the mask directly at the target sparsity. Nested coarse pruning instead repeatedly recalibrates LCAM on the currently sparse policy and removes only additional connections. The full method further activates action-aware fine-grained allocation after the detected sparsity knee.

Nested coarse pruning improves over one-shot pruning at $40\%$ and $50\%$. At $60\%$, both coarse-only variants collapse, whereas the full coarse-to-fine procedure retains $68.5\%$ success. This indicates that recursive recalibration is useful before the sensitive regime, while the fine stage is important once additional pruning induces a sharp increase in offline action distortion.

\begin{table}[t]
\centering
\caption{Pruning-procedure ablation on LIBERO-Object (\%).
All variants use the full LCAM score.}
\label{tab:procedure_ablation}
\scriptsize
\setlength{\tabcolsep}{4.4pt}
\renewcommand{\arraystretch}{1.08}
\begin{tabular}{lccc}
\hline
Method & $40\%$ & $50\%$ & $60\%$ \\
\hline
One-shot Native          & 69.5 & 56.5 & 0.0 \\
Nested Coarse LCAM       & 72.5 & 68.5 & 0.0 \\
\textbf{Full LCAM}       & \textbf{72.5} & \textbf{68.5} & \textbf{68.5} \\
\hline
\end{tabular}
\end{table}

\paragraph{Effect of the LCAM importance criterion}
For score ablation, all variants use the same nested coarse pruning procedure and pruning configuration; only the LCAM connection score is modified. \textit{w/o Action-Loss Sensitivity} removes both the loss-conditioned activation modulation and the native layer-level loss scale. \textit{w/o Activation Moments} removes the first- and second-order
activation-moment terms jointly, while \textit{w/o Row-Normalized Weight Contribution} removes the row-normalized weight factor.

Table~\ref{tab:score_ablation} shows that the three signals play complementary roles. Removing action-loss sensitivity causes a sharp performance drop, indicating that activation evidence alone cannot distinguish connections whose perturbation is consequential to the action prediction objective. Removing activation moments also substantially
degrades performance, suggesting that loss sensitivity should be evaluated in the context of how consistently a connection is activated by the demonstrations. Finally, removing the row-normalized weight contribution leads to near-complete failure, suggesting that row-wise normalization helps prevent the score from being dominated by raw activation or gradient magnitudes across heterogeneous output channels. Together, these results support the use of row-normalized weight contribution, activation evidence, and action-loss sensitivity as complementary components of the LCAM score.

\begin{table}[t]
\centering
\caption{LCAM score ablation on LIBERO-Object (\%).
All variants use nested coarse pruning.}
\label{tab:score_ablation}
\scriptsize
\setlength{\tabcolsep}{3.8pt}
\renewcommand{\arraystretch}{1.08}
\begin{tabular}{lcc}
\hline
Score variant & $40\%$ & $50\%$ \\
\hline
Full LCAM                         & \textbf{72.5} & \textbf{68.5} \\
w/o Action-Loss Sensitivity       & 17.5 & 1.0 \\
w/o Activation Moments            & 59.0 & 0.0 \\
w/o Row-Normalized Weight  & 0.0  & 0.0 \\
\hline
\end{tabular}
\end{table}

\section{CONCLUSIONS}
\label{sec:conclusion}

We introduce LCAM, a training-free post-training method for unstructured pruning of robotic manipulation policies. LCAM uses loss-conditioned activation evidence to rank connections and applies self-recursive coarse-to-fine pruning to construct nested masks from offline demonstrations alone, without recovery training or simulator rollouts. Across three LIBERO suites with MATE policies, LCAM retains substantial task success at high pruning ratios where several compared baselines sharply degrade or collapse. The evaluation on OpenVLA further shows that LCAM achieves the highest success among the compared methods at $50\%$ and $60\%$ pruning; at $50\%$ pruning, it retains over $90\%$ of the dense policy's success rate. Overall, these results support loss-conditioned, activation-aware importance estimation as a promising training-free direction for preserving robotic policy behavior under high unstructured sparsity.

LCAM also has limitations. Offline action distortion is a surrogate for closed-loop task success and may not fully capture long-horizon interaction effects at extreme sparsity. Moreover, the current evaluation covers three MATE suites and one OpenVLA setting, so broader generalization across policy architectures, robot platforms, and real-world tasks remains to be established. Future work may explore offline pruning criteria that better reflect long-horizon action effects and evaluate LCAM across a wider range of robotic policy families and manipulation settings.


%%%%%%%%%%%%%%%%%%%%%%%%%%%%%%%%%%%%%%%%%%%%%%%%%%%%%%%%%%%%%%%%%%%%%%%%%%%%%%%%

%%%%%%%%%%%%%%%%%%%%%%%%%%%%%%%%%%%%%%%%%%%%%%%%%%%%%%%%%%%%%%%%%%%%%%%%%%%%%%%%

%%%%%%%%%%%%%%%%%%%%%%%%%%%%%%%%%%%%%%%%%%%%%%%%%%%%%%%%%%%%%%%%%%%%%%%%%%%%%%%%
\section*{APPENDIX}

\subsection{Pruning Data Protocol}

For each policy backbone, we distinguish calibration data, held-out allocation data, and final evaluation data. The calibration set $\mathcal{D}_{\mathrm{cal}}$ is used to estimate LCAM statistics, whereas the disjoint allocation set $\mathcal{D}_{\mathrm{val}}$ is used only to compute offline action distortion for automatic knee detection and fine-stage candidate-mask selection. Final evaluation data are reserved exclusively for reporting policy performance and are never accessed during mask construction. MATE uses its native dataset splits, whereas OpenVLA uses the corresponding RLDS-formatted LIBERO data. Within each backbone, all compared methods use the same fixed pruning seed when constructing the calibration and allocation streams.

\subsection{MATE Pruning Implementation}

For MATE, we apply post-training unstructured weight pruning to all linear layers except those in the frozen track branch, including the action-prediction head. To avoid fully disconnecting individual input or output features, we use the same row- and column-retention safeguards for all compared pruning methods.

The native MATE action-prediction objective is the mean-squared error
between the predicted continuous actions and the demonstrated actions:
\begin{equation}
\mathcal{L}_{\mathrm{act}}^{\mathrm{MATE}}
=
\operatorname{MSE}
\left(
\hat{\mathbf{a}},\mathbf{a}
\right).
\end{equation}
This loss is evaluated on calibration batches only to obtain the
gradients used by LCAM.

LCAM statistics are estimated from 16 calibration batches, whereas the offline action distortion $D(\mathbf{M})$ defined in Eq.~\eqref{eq:action_distortion} is evaluated on 16 batches from the held-out allocation set $\mathcal{D}_{\mathrm{val}}$. Data augmentation is disabled for both streams during pruning. Before mask construction, we cache the dense-policy outputs $\pi_\theta(\mathbf{o})$ for $\mathbf{o}\in\mathcal{D}_{\mathrm{val}}$ and use them as the common reference for evaluating all candidate masks. The per-dimension standard-deviation vector $\boldsymbol{\sigma}$ in Eq.~\eqref{eq:action_distortion} is estimated from the calibration stream, and we set the stability floor to $\tau=0.05$. During the second LCAM pass, at most 2048 action-supervised positions are used for each layer call when accumulating joint moments.

The automatic controller follows the validation-controlled coarse-to-fine procedure in Section~\ref{subsec:coarse_to_fine}. It uses global sparsity increments of 10\% during the coarse stage and 1\% during the fine stage. The knee test is activated after two prior coarse marginal distortions are available, with the sensitivity multiplier set to $\beta=2.0$. LCAM statistics are recomputed for each coarse candidate and at the beginning of every fine allocation round. The controller retains a CPU-side snapshot of the most recently accepted coarse model. If a coarse candidate triggers the knee criterion, it is rejected and the model is restored to the latest accepted snapshot, which defines the boundary for fine allocation.

At each fine allocation step, equal-size candidate pruning blocks are formed for all feasible functional groups. These groups correspond to the action-prediction head, spatial-attention modules, spatial MLP modules, temporal Transformer modules, and the remaining encoder modules. Connections within each candidate group are selected using the current LCAM-native criterion under the same retention safeguards. The candidate with the smallest increase in offline action distortion per newly pruned connection is accepted and applied as the next sparsity block. Neither recovery training nor simulator rollouts are used during mask construction.

\subsection{OpenVLA Pruning Implementation}

For OpenVLA, we use the action-token prediction loss as $\mathcal{L}_{\mathrm{act}}$ when computing LCAM scores for the Llama Transformer projection matrices; the visual encoder is left unchanged. LCAM statistics are estimated from 16 calibration samples. Each sample is constructed from one transition selected from an RLDS training episode.

For OpenVLA, the native action-prediction objective is the mean
negative log-likelihood over the supervised action-token positions:
\begin{equation}
\mathcal{L}_{\mathrm{act}}^{\mathrm{OpenVLA}}
=
-\operatorname{mean}_{t\in\mathcal{T}_{\mathrm{sup}}}
\log p_{\theta}
\left(
y_t\mid\mathbf{o},y_{<t}
\right),
\end{equation}
where $\mathcal{T}_{\mathrm{sup}}$ denotes the supervised positions in
the action-token target sequence. This loss is used only to obtain LCAM
gradients during calibration.

The selected Llama projection matrices contain approximately 6.4 billion weights in the OpenVLA-7B model. Storing both joint-moment accumulators for all target layers simultaneously is therefore impractical. We use a memory-bounded implementation in which the first pass collects the output-gradient RMS for every target layer. During the second pass, Transformer blocks are processed in chunks of two blocks. For a layer call containing more than 2048 token positions, a deterministic subset of 2048 positions is used for moment accumulation.

OpenVLA uses the same validation-controlled coarse-to-fine controller
as MATE, with global sparsity increments of 10\% and 1\% in the coarse
and fine stages, respectively. The knee test is activated after two
preceding coarse marginal distortions have been observed, with
sensitivity multiplier $\beta=2.0$.

For each block chunk, masks are computed from the same parent checkpoint
and stored temporarily as bit-packed files. Model weights remain
unchanged until all chunks have completed calibration, after which the
newly selected masks are applied together. Thus, all layers within one
nested pruning stage are scored against the same parent model. Existing
zeros are preserved across successive stages, yielding nested masks. No
recovery training or simulator rollouts are used during pruning.

% \section*{ACKNOWLEDGMENT}
% This work was supported by National Natural Science Foundation of China (NSFC) under Grant Nos. 62376283 and 62531026; by the Fundamental and Interdisciplinary Disciplines Breakthrough Plan of the Ministry of Education of China under Grant JYB2025XDXM110; and by the Innovation Research Foundation of National University of Defense Technology under Grant JS2023-03.

\bibliographystyle{IEEEtran}
\bibliography{references}

\end{document}